\documentclass[letterpaper, 10 pt, conference]{ieeeconf}
\IEEEoverridecommandlockouts
\usepackage{multicol}
\usepackage{footmisc}
\usepackage{amsmath,amssymb,amsfonts}
\usepackage{algorithmic}
\usepackage{graphicx}
\usepackage{float}
\usepackage[ruled,linesnumbered,vlined]{algorithm2e}
\usepackage{verbatim}
\usepackage{tikz}
\usepackage{pgfplots}

\usetikzlibrary{shapes.geometric}
\usetikzlibrary{patterns}
\usetikzlibrary{calc}
\usepackage{subcaption}
\usepackage{caption}
\usepackage{url}

\usepackage[section]{placeins}
\pgfplotsset{compat=1.16}
\SetArgSty{textnormal}
\usepackage{array}
\newif\ifanonymize

\title{\LARGE \bf
  Kinodynamic Motion Planning for Mobile Robot Navigation across Inconsistent World Models 
}

\ifanonymize
  \author{
    Author list anonymized for peer review$^1$%
    \thanks{\newline\newline\newline*Research sponsors anonymized for peer review\newline\newline\newline}
    \thanks{\newline\newline$^1$Author affiliations anonymized for peer review\newline\newline}
  }
\else
  \author{Eric R. Damm$^{1}$ and Thomas M. Howard$^{1}$
  \thanks{*Research was sponsored by the DEVCOM Army Research Laboratory (ARL) and was accomplished under Cooperative Agreement Number W911NF-20-2-0106. The views and conclusions contained in this document are those of the authors and should not be interpreted as representing the official policies, either expressed or implied, of the Army Research Laboratory or the U.S. Government. The U.S. Government is authorized to reproduce and distribute reprints for Government purposes notwithstanding any copyright notation herein.}
  \thanks{$^{1}$Eric R. Damm and Thomas M. Howard are with the Robotics and Artificial Intelligence Laboratory, Hajim School of Engineering and Applied Sciences,
    University of Rochester, Rochester, NY, USA
          {\tt\small edamm@ur.rochester.edu}}%
  }
\fi

\begin{document}
\maketitle
\thispagestyle{empty}
\pagestyle{empty}
\begin{abstract}
Mobile ground robots lacking prior knowledge of an environment must rely on sensor data to develop a model of their surroundings.
In these scenarios, consistent identification of obstacles and terrain features can be difficult due to noise and algorithmic shortcomings, which can make it difficult for motion planning systems to generate safe motions.
One particular difficulty to overcome is when regions of the cost map switch between being marked as obstacles and free space through successive planning cycles.
One potential solution to this, which we refer to as Valid in Every Hypothesis (VEH), is for the planning system to plan motions that are guaranteed to be safe through a history of world models.
Another approach is to track a history of world models, and adjust node costs according to the potential penalty of needing to reroute around previously hazardous areas.
This work discusses three major iterations on this idea.
The first iteration, called Per-Edge Hypothesis (PEH), invokes a sub-search for every node expansion that crosses through a divergence point in the world models.
The second and third iterations, called Goal-Edge Hypothesis (GEH) and Goal-Edge Graph Revision Hypothesis (GEGRH) respectively, defer the sub-search until after an edge expands into the goal region.
GEGRH uses an additional step to revise the graph based on divergent nodes in each world.
Initial results showed that, although PEH and GEH find more optimistic solutions than VEH, they are unable to generate solutions in less than one-second, which exceeds our requirements for field deployment.
Analysis of results from a field experiment in an unstructured, off-road environment on a Clearpath Robotics Warthog Unmanned Ground Vehicle (UGV) indicate that GEGRH finds lower cost trajectories and has faster average planning times than VEH. 
Compared to single-hypothesis (SH) search, where only the latest world model is considered, GEGRH generates more conservative plans with a small increase in average planning time.
\end{abstract}

\section{Introduction}
Robots navigating through unstructured, partially observed environments are often tasked with generating their own models of the world during traversal.
These models are used to inform motion planners of environmental hazards, so they can generate safe trajectories.
Traditionally, motion planners that operate to or beyond the perception horizon build a graph in which edge costs are influenced by the most recent world map.
Search algorithms such as A$^*$ \cite{HartASTAR} or Anytime Repairing A$^*$ (ARA$^*$) \cite{likhachev2003ara} are then used to generate feasible motions.
By relying on the most recent world representation, the planner attempts to ensure that its output reflects the most accurate approximation of the environment.

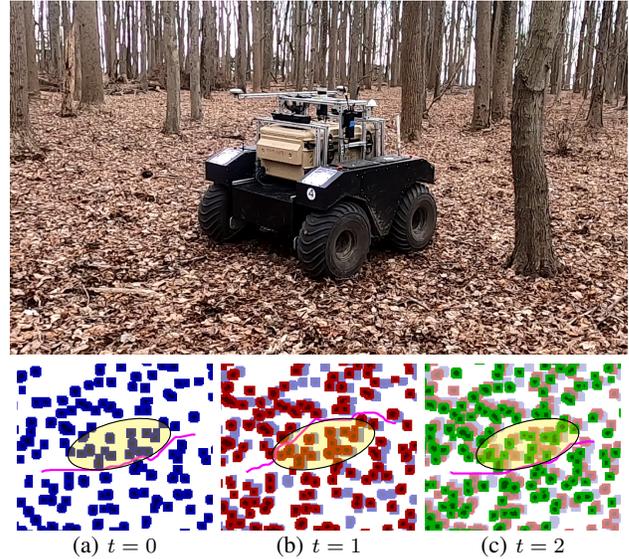
\begin{figure}
	\centering
	\begin{subfigure}{.95\linewidth}
		\includegraphics[width=\linewidth]{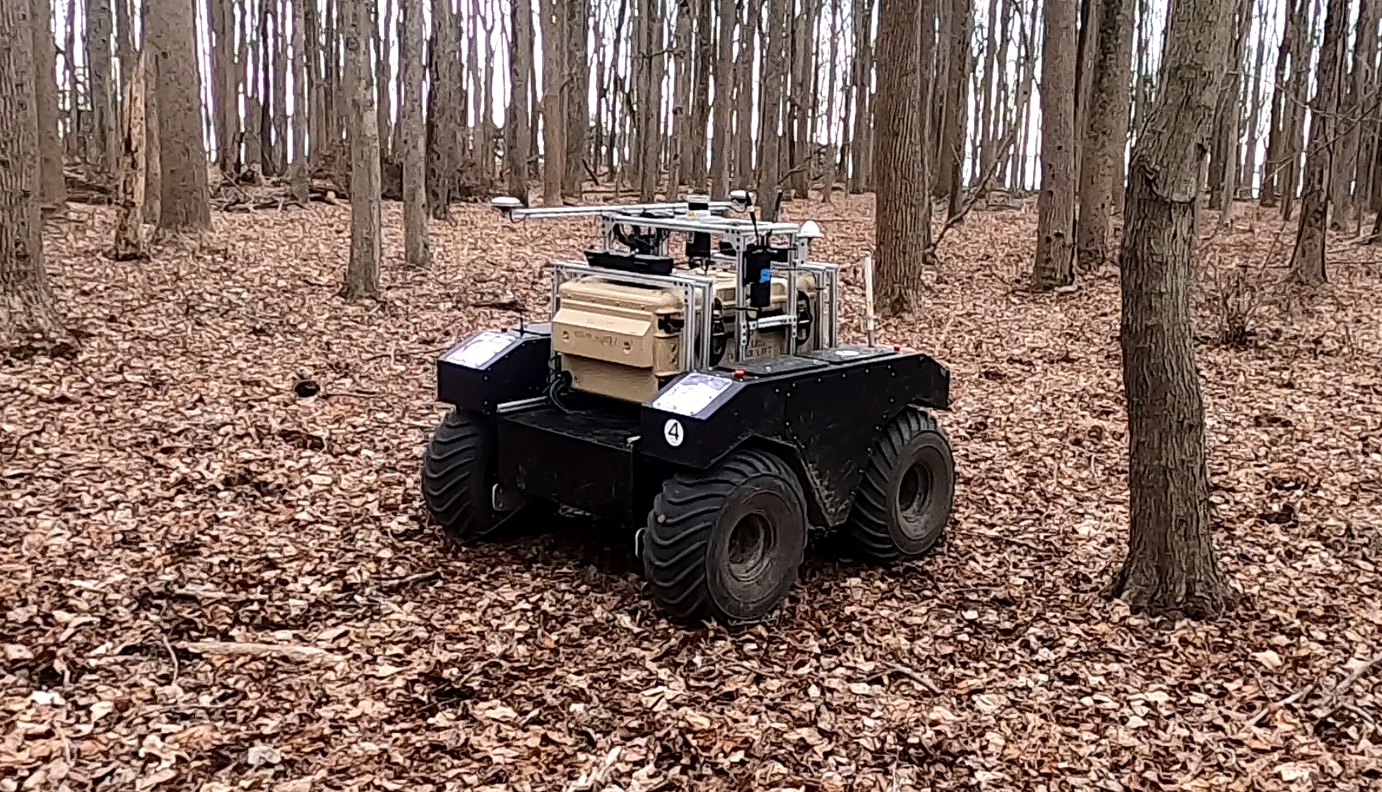}
		\vspace{-.3cm}
	\end{subfigure}
	\begin{subfigure}{0.3\linewidth}
		\begin{tikzpicture}
			\node[anchor=south west,inner sep=0] at (0,0) { \includegraphics[width=\linewidth,trim={11.5cm 9cm .5cm 5cm},clip]{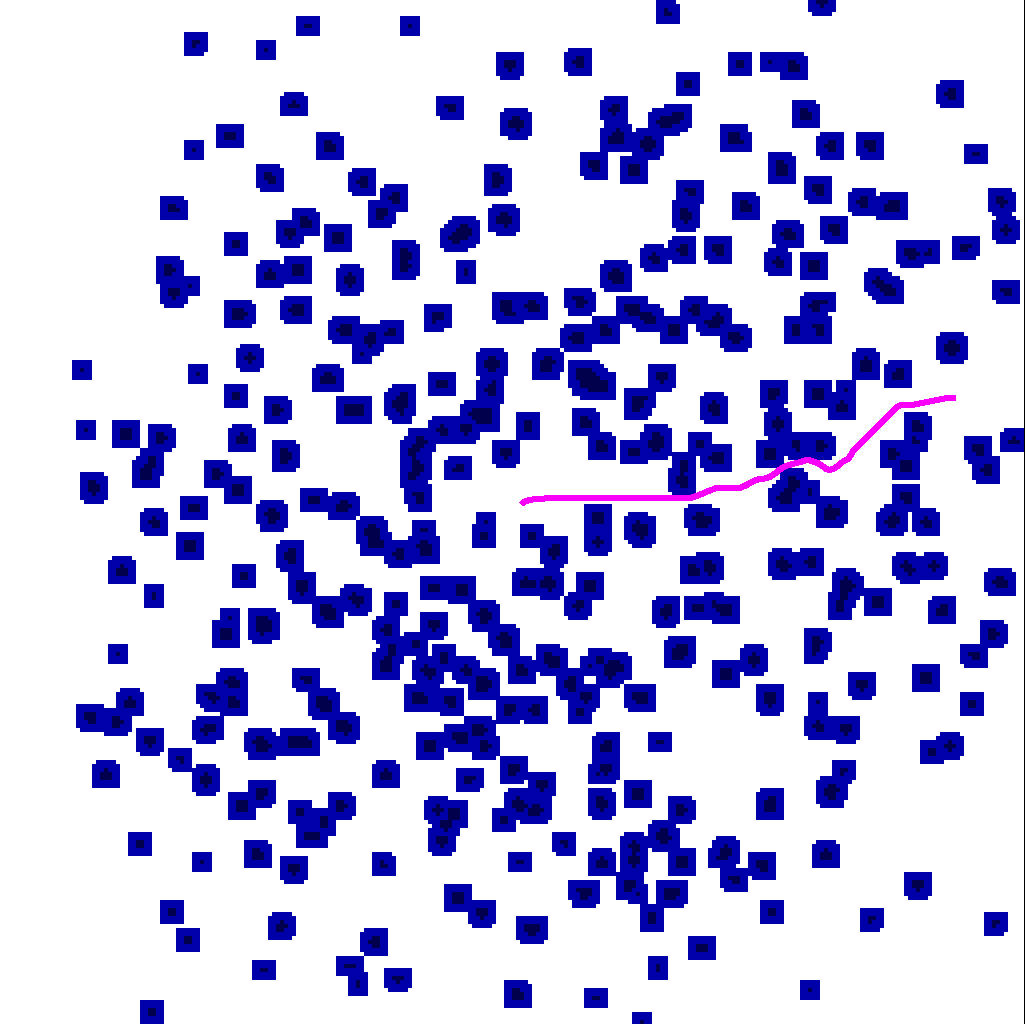} };
			\draw[rotate around={15:(1.35,1.2)}, black, thin] (1.35,1.15) ellipse[x radius=0.7, y radius=0.3];
			\draw[rotate around={15:(1.35,1.2)}, fill=yellow, opacity=0.3] (1.35,1.15) ellipse[x radius=0.7, y radius=0.3];
		\end{tikzpicture}
		\caption{$t=0$}
	\end{subfigure}
	\begin{subfigure}{0.3\linewidth}
		\begin{tikzpicture}
			\node[anchor=south west,inner sep=0] at (0,0) { \includegraphics[width=\linewidth,trim={11.5cm 9cm .5cm 5cm},clip]{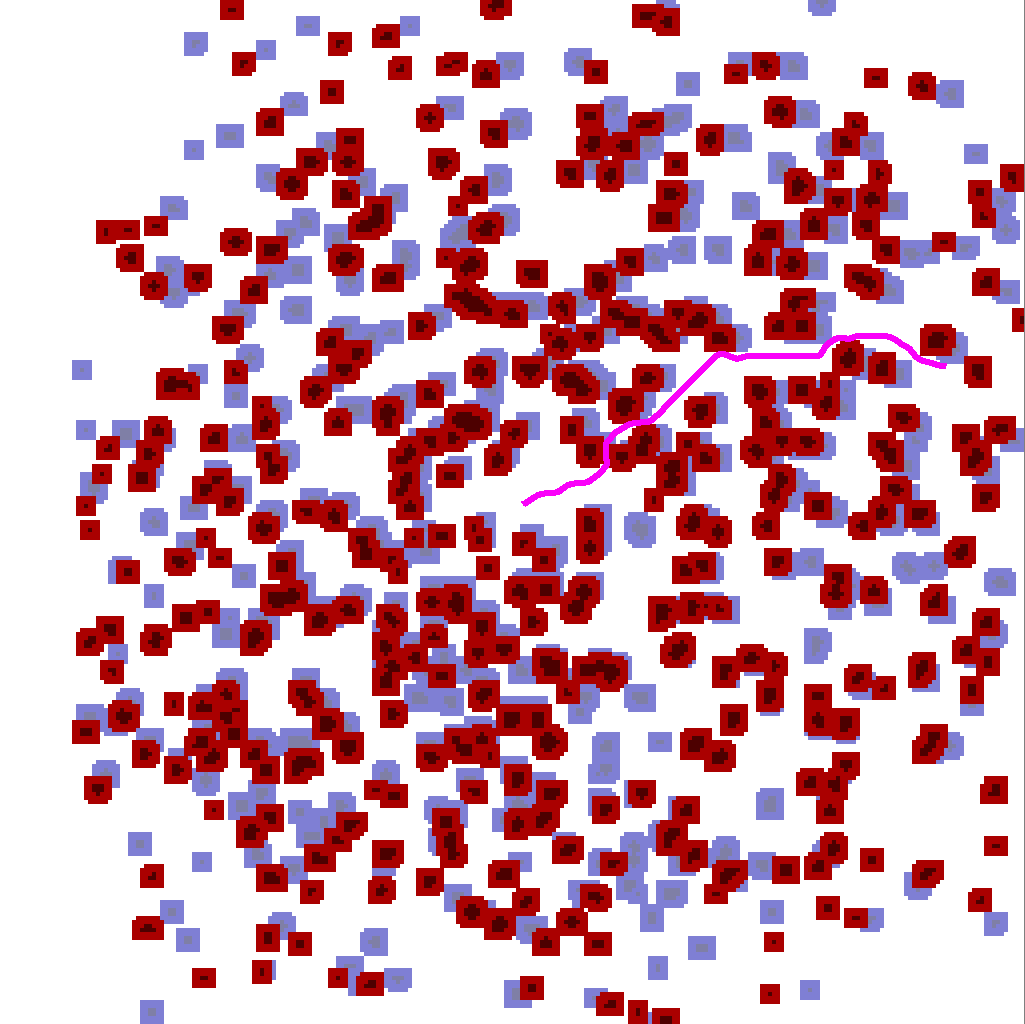} };
			\draw[rotate around={15:(1.35,1.2)}, black, thin] (1.35,1.15) ellipse[x radius=0.7, y radius=0.3];
			\draw[rotate around={15:(1.35,1.2)}, fill=yellow, opacity=0.3] (1.35,1.15) ellipse[x radius=0.7, y radius=0.3];
		\end{tikzpicture}
		\caption{$t=1$}
	\end{subfigure}
	\begin{subfigure}{0.3\linewidth}
		\begin{tikzpicture}
			\node[anchor=south west,inner sep=0] at (0,0) { \includegraphics[width=\linewidth,trim={11.5cm 9cm .5cm 5cm},clip]{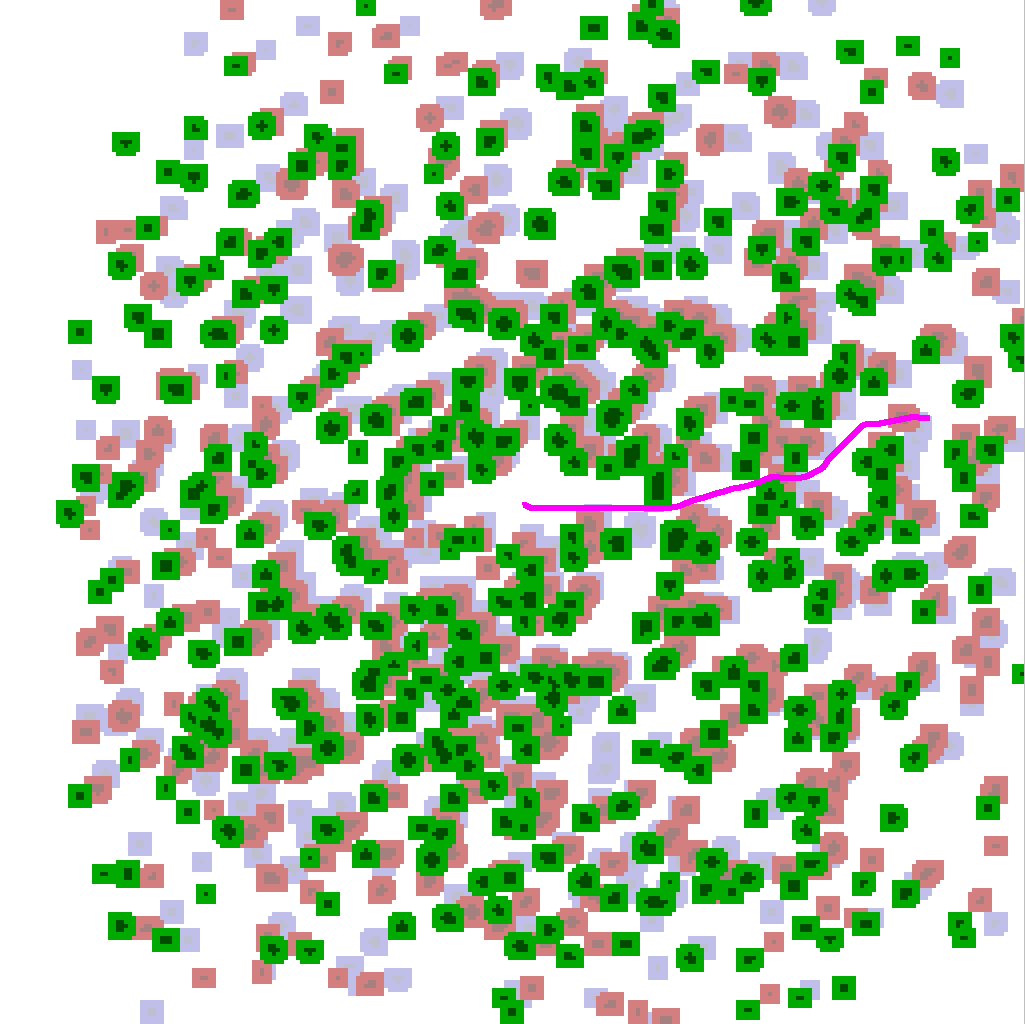} };
			\draw[rotate around={15:(1.35,1.2)}, black, thin] (1.35,1.15) ellipse[x radius=0.7, y radius=0.3];
			\draw[rotate around={15:(1.35,1.2)}, fill=yellow, opacity=0.3] (1.35,1.15) ellipse[x radius=0.7, y radius=0.3];
		\end{tikzpicture}
		\caption{$t=2$}
	\end{subfigure}
	\caption{Top: UGV in an unstructured, off-road environment. Bottom: Three successive planning problems, with the colored cost map overlaid on a faded image of the previous. Relying on the most recent cost map causes the planned trajectory (magenta) to oscillate around the central obstacles (shaded yellow, circled in black).}
	\label{fig:overlaidWorlds}
\vspace{-.2cm}
\end{figure}

In practice, these maps are subject to uncertainty from sensor noise, occlusions, state estimation errors, and incomplete observations, particularly in complex and partially observable environments \cite{lazanas1995motion}.
This uncertainty results in inconsistencies between successive world models, undermining the assumption that the most recent map is the most accurate.
The differences in these maps lead to inconsistent plan generation from deterministic motion planners.
This effect was highlighted during a field test on a Clearpath Robotics Unmanned Ground Vehicle (UGV).
Figure \ref{fig:overlaidWorlds} is an illustration from the experiment that shows the variation in perception output by overlaying three successive world maps \cite{meng2023terrainnetvisualmodelingcomplex}.
Even in near-field regions where sensor confidence is typically higher, inconsistencies exist in what is classified as lethal in the cost maps. 
These inconsistencies introduce deviations between successive planner outputs, leading to undesirable behaviors such as abrupt trajectory changes or oscillations around obstacles.
This oscillatory behavior, also shown in Figure \ref{fig:overlaidWorlds}, is described in a topological sense in \cite{HAEASL}, where trajectories oscillate back and forth due to changes in the cost map.
Such oscillations arise when the planner repeatedly generates trajectories that diverge from previous plans, leading to instability in robot motion.

The planner used for this work leverages the Kinodynamic Efficiently Adaptive State Lattice (KEASL) search space with heuristic based search \cite{KEASL}.
Figure \ref{fig:overlaidWorlds} shows the oscillatory plans generated by KEASL.
This behavior, where alternating, topologically distinct trajectories (shown in magenta) are generated through the most recent world map, is due to differences in the successive inputs to the motion planner.
The deterministic properties of KEASL mean the outputs are predictable, and the behaviors can be explained algorithmically.
Additionally, the discrete environment representations mean there is a consistent output with a consistent input.
The oscillatory problems arise because the opposite can be true, where noisy inputs tend to result in noisy outputs.
Stochastic processes can take the uncertainty into account, but often require training models or carefully tuning parameters.
Partially Observable Markov Decision Processes (POMDP), for example, can handle uncertainty in observations, but require updating a policy for each new map.

\begin{figure}[ht]
	\centering
	\begin{subfigure}{0.3\linewidth}
		\begin{tikzpicture}
			\node[anchor=south west,inner sep=0] at (0,0) { \includegraphics[width=\linewidth,trim={11.5cm 9cm .5cm 5cm},clip]{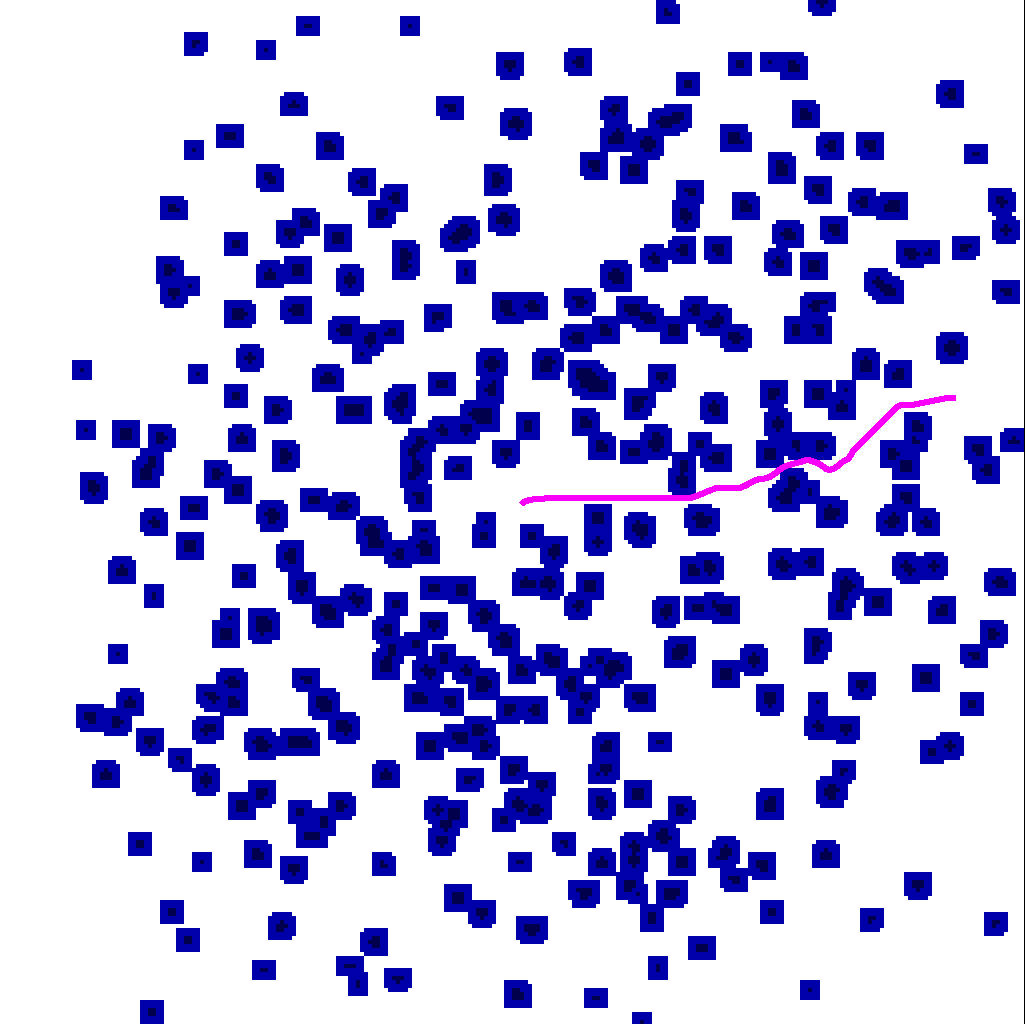} };
			\draw[rotate around={15:(1.35,1.2)}, black, thin] (1.35,1.15) ellipse[x radius=0.7, y radius=0.3];
			\draw[rotate around={15:(1.35,1.2)}, fill=yellow, opacity=0.3] (1.35,1.15) ellipse[x radius=0.7, y radius=0.3];
		\end{tikzpicture}
		\caption{$t=0$}
	\end{subfigure}
	\begin{subfigure}{0.3\linewidth}
		\begin{tikzpicture}
			\node[anchor=south west,inner sep=0] at (0,0) { \includegraphics[width=\linewidth,trim={11.5cm 9cm .5cm 5cm},clip]{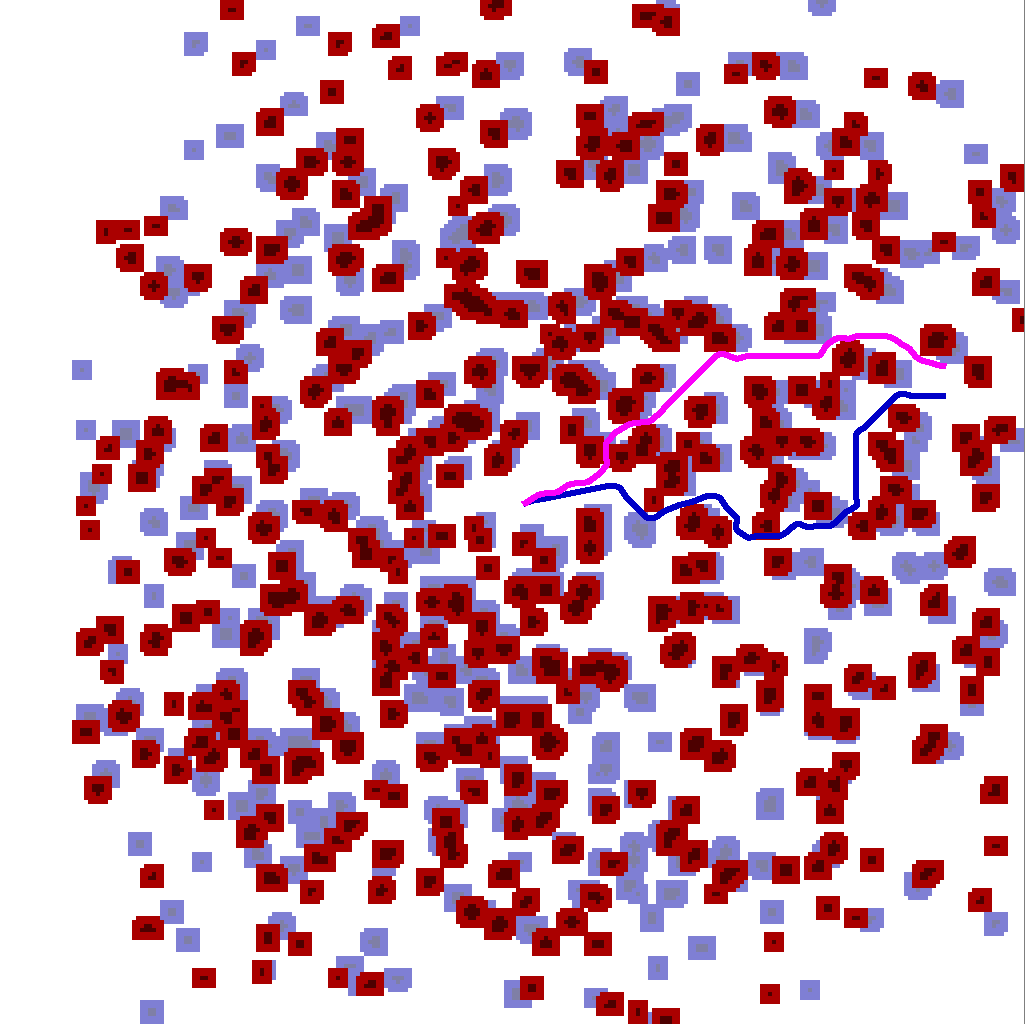} };
			\draw[rotate around={15:(1.35,1.2)}, black, thin] (1.35,1.15) ellipse[x radius=0.7, y radius=0.3];
			\draw[rotate around={15:(1.35,1.2)}, fill=yellow, opacity=0.3] (1.35,1.15) ellipse[x radius=0.7, y radius=0.3];
		\end{tikzpicture}
		\caption{$t=1$}
	\end{subfigure}
	\begin{subfigure}{0.3\linewidth}
		\begin{tikzpicture}
			\node[anchor=south west,inner sep=0] at (0,0) { \includegraphics[width=\linewidth,trim={11.5cm 9cm .5cm 5cm},clip]{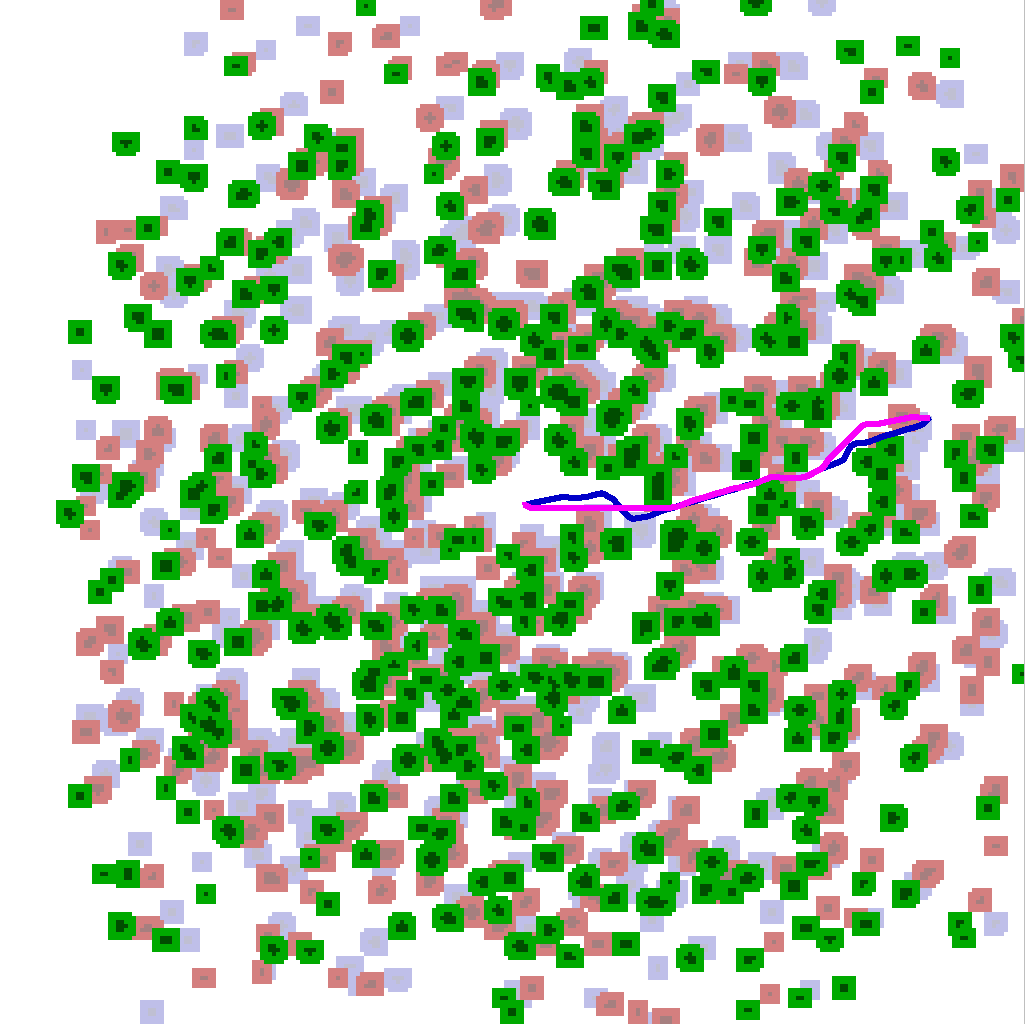} };
			\draw[rotate around={15:(1.35,1.2)}, black, thin] (1.35,1.15) ellipse[x radius=0.7, y radius=0.3];
			\draw[rotate around={15:(1.35,1.2)}, fill=yellow, opacity=0.3] (1.35,1.15) ellipse[x radius=0.7, y radius=0.3];
		\end{tikzpicture}
		\caption{$t=2$}
	\end{subfigure}
	\caption{Three successive planning problems, with the colored cost map overlaid on a faded image of the previous. Using only the most recent cost map causes the planned trajectory (magenta) to oscillate around the central clustering of obstacles (shaded yellow, circled in black). The multi-hypothesis plan (blue) is consistently to one side of the cluster. Note that at $t=0$ the trajectories are identical because only one world hypothesis exists at this point.}
	\label{fig:planOscillations2}
\end{figure}
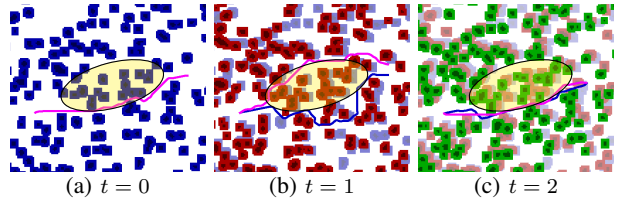

Since mapping uncertainty can significantly impact planning performance, we propose a more robust approach to edge cost computation that uses multiple temporally sampled environment maps.
By considering prior perception decisions and integrating them into the search process, the planner can reason about prior world observations and make more informed decisions.
We characterize each successive world map as an evolving hypothesis about the true state of the environment.
Divergence points between these hypotheses correspond to regions where the perception system's interpretation of the environment has changed, often leading to inconsistencies in the plans generated by motion planning algorithms.
Each map in Figure \ref{fig:overlaidWorlds} can be viewed as the accumulation of prior hypotheses, each modifying previous interpretations of the environment.
In this work, the most recent map is treated as the primary hypothesis, where there cannot be any collisions in the final solution.
The additional hypotheses are used to guide the search process.
%
Figure \ref{fig:planOscillations2} shows the output from our multi-hypothesis planning methodology with the same environment map representations as in Figure \ref{fig:overlaidWorlds}.
The Single-Hypothesis (SH) plan (magenta) oscillates around a central clustering of obstacles, while the multi-hypothesis plan (blue) remains consistently to one side.
While our approach does not completely eliminate this behavior, it mitigates the impact by making decisions with consideration of prior world observations.

This work presents two main contributions aimed at addressing the challenges posed by perceptual uncertainty in the environment representation:
\begin{enumerate}
\item A search algorithm for mobile robot motion planning in off-road environments that considers multiple temporally sampled world model hypotheses when computing edge costs.
\item An analysis of the algorithm's performance with experiments conducted on data collected from field-testing on a UGV in an unstructured, off-road environment.
\end{enumerate}
\section{Related Work}
There is a large body of research addressing the challenge of overcoming environmental uncertainty in motion planning.
However, most of the work is in mitigating the impacts of uncertainty in localization and action response.
There is an open research area regarding the gap between inconsistencies in environment map representations and the results of deterministic planners.

The Markov Decision Process (MDP) framework has been extensively used to model decision-making under uncertainty when the system's current state is fully observable \cite{puterman2014markov}.
In an MDP, a policy is computed to maximize expected rewards over time, accounting for probabilistic transitions between states caused by the agent's actions.
MDPs have been widely applied in motion planning tasks where action outcomes are stochastic but the environment can be reliably sensed, allowing planners to handle uncertainty without requiring a full belief state representation.

The POMDP is an extension of the MDP, and is used for handling uncertainty in world models \cite{POMDP,simmons1995probabilistic}.
POMDPs account for incomplete or noisy observations by maintaining a belief over possible system states and updating it based on incoming sensor data.
To optimize decision-making in partially observed environments, value iteration methods have been extensively explored to determine optimal policies \cite{porta2006point,kurniawati2008sarsop}.
These approaches enable a robot to reason about the probabilistic outcomes of actions, but require predictions of the uncertainty of the world. 
While POMDP frameworks are robust to observation noise, they are often computationally infeasible for field deployment due to the high dimensionality of long-range, kinodynamic motion planning \cite{roy1999coastal,kaelbling1998planning}.
Policy computation is costly, and in noisy and partially observed environments, the policy may need frequent updates to reflect changes in the map.

Another effective approach to managing uncertainty is belief space planning, where motions are generated based on the robot's predicted ability to localize and reduce uncertainty at each decision point.
Belief space planning considers a probability distribution over the possible states of the environment, enabling the robot to choose actions that maximize information gain or minimize uncertainty \cite{bry2011rapidly,gonzalez2014state,pairet2021online}.
This method is particularly useful in situations where maintaining localization accuracy is important, at the expense of traditionally suboptimal plans.

In dynamic and partially observed environments, considering multiple possible outcomes can significantly improve motion planning performance.
%
%
One such approach solves multiple plans conditioned on the possible actions of dynamic agents and collapses them into a final solution once the true behavior is observed \cite{davis2018multiworld}.
Similarly, \cite{gong2011multi} propose a motion model for tracking and predicting the behavior of multiple objects, generating plausible plans for each object in the scene to ensure safe and effective navigation.
Additionally, work by \cite{miura2002probabilistic} demonstrates advancements in predicting potential obstacle trajectories, further refining the application of probabilistic models to motion planning in dynamic scenarios.

Another class of approaches aims to eliminate uncertainty by filling in missing or ambiguous information at the perception level.
One way to do this is by incorporating hysteresis into the traversability estimates like in \cite{urmson2007tartan}, where the cost map cells with the highest cost were only removed from the map after a certain amount time.
Another way to handle the uncertainty is by using predictive methods which fill in uncertain areas.
These methods use prior knowledge or learned models to infer the structure of the environment where sensor data is sparse or noisy.
Work by \cite{meng2023terrainnetvisualmodelingcomplex} uses a model to generate world models in unstructured environments, and \cite{elhafsi2020map} demonstrates an approach where hallway shapes are predicted and reconstructed during simulated robot traversal, allowing the robot to make informed navigation decisions in partially mapped environments.
Although such predictive techniques enhance the planner's confidence in unobserved regions, overconfidence by the models can lead to inconsistent classification of hazardous regions.

Other work has addressed the problem of planning with inconsistent world models by generating individual motions over a posterior of map observations, and selecting the final trajectory based on the lowest aggregate cost of those motions in each world sample \cite{schmittle2024multi}.
While this approach does consider multiple map representations during path selection, it guarantees relative optimality in only one world.
In contrast, our method considers multiple maps during search to generate an optimal motion based on the potential cost of rerouting around inconsistently identified hazards.

Our work differs from prior methods by encoding inconsistent environment predictions into a deterministic graph, where edge costs are computed from temporally sampled world hypotheses.
This enables deterministic, kinodynamic planners like KEASL to maintain efficient and accurate search without relying on parameter tuning or model training to compensate for environmental noise.
Additionally, we differentiate between modeling uncertain regions as hazards and assigning them costs that reflect the potential risk of incorrect assumptions.
%
%
\section{Technical Approach}\label{sec:technicalApproach}
The KEASL search space was developed to perform kinodynamic motion planning with consistent velocity constraints.
With traditional KEASL (herein referred to as Single-Hypothesis (SH) planning), a library of motion primitives is used to generate feasible expansions.
Each expansion is validated against the most recent world map by first checking for obstacle collisions and then applying velocity constraints.
Expansions that satisfy both conditions are added to the recombinant graph, where nodes maintain a history of expansions leading back to the start.
For context, the f-cost refers to the sum of the heuristic estimate to the goal and the total estimated time it would take to reach a node.
The open list contains nodes available for expansion, while the closed list contains nodes that have already been expanded.

To develop the contributions in this paper, we explored four different methods of searching over a recombinant, multi-world-hypothesis search space.
The first method, and our baseline, ensures motion viability across all world representations, expanding an edge only if it is collision-free in every hypothesis.
The second checks for motion validity on a per-expansion basis and invokes a sub-search to reroute edges around obstacles in corresponding worlds.
The third method implements a ``lazy search'' method \cite{mandalika2019generalized} by deferring the acknowledgement of multiple hypotheticals until after an edge expands to the goal.
After such an expansion, the candidate solution is checked for collisions in each hypothesis and rerouted from the corresponding divergence based on the considered world.
The final method builds on the third method by incorporating a graph-revision step after performing the rerouting in each world. 
In methods two, three, and four, edges are expanded if they are valid in at least one of the worlds.
It is not until the final step in each process that the solution must be valid in the most recent hypothesized world.

\subsection{Multi-World-Hypothesis Search Space}\label{sec:searchSpace}
We build upon the KEASL search space by considering multiple hypothetical world models when computing edge costs.
Each node in the graph has a backpointer, which is a reference to its parent, and a list of the expansions from the prior nodes in its history.
For this work, the backpointers remain consistent with KEASL, but we reformat the node's expansion history data structure to include one history per world hypothesis.
This allows expansions to occur through cost map cells that contain obstacles in some hypotheses, and do not in others, while tracking the points of divergence in the node histories.
By formulating the search space this way, we can detect and bridge through inconsistently observed regions of the world.
%
%

\subsection{Valid in Every Hypothesis (VEH)}
The first approach ensures that expanded edges are valid in every hypothetical world in the graph.
This is done by checking expansion validity against every world representation during search.
Although searching in this way ensures obstacle consideration in inconsistent world models, the generated plans are often overly conservative.
Pseudocode for VEH is shown in Algorithm \ref{alg:VEH}.
\begin{algorithm}[htb]
\footnotesize
\caption{VEH}\label{alg:VEH}
\begin{algorithmic}[1]
	\STATE $OPEN \gets$ {start node}
	\STATE $CLOSED \gets \emptyset$
	\WHILE{$OPEN$ not empty}
			\STATE $n \gets$ $OPEN$.pop()
			\STATE $CLOSED \gets n$
			\IF{$n$ is goal}
					\RETURN extractSolution($n$)
			\ENDIF
			\FORALL{$n'$ in Expand($n$)}
					\IF{$n'$ is valid in \textbf{every} hypothesis}
							\STATE UpdateCost($n$, $n'$)
							\STATE $OPEN$.insert($n'$)
					\ENDIF
			\ENDFOR
	\ENDWHILE
\end{algorithmic}
\end{algorithm}

\begin{figure}[htb]
	\vspace{-.4cm}
	\centering
	\begin{tikzpicture}
		\node[anchor=south west,inner sep=0] at (0,0) {\includegraphics[height=.85\linewidth,angle=90]{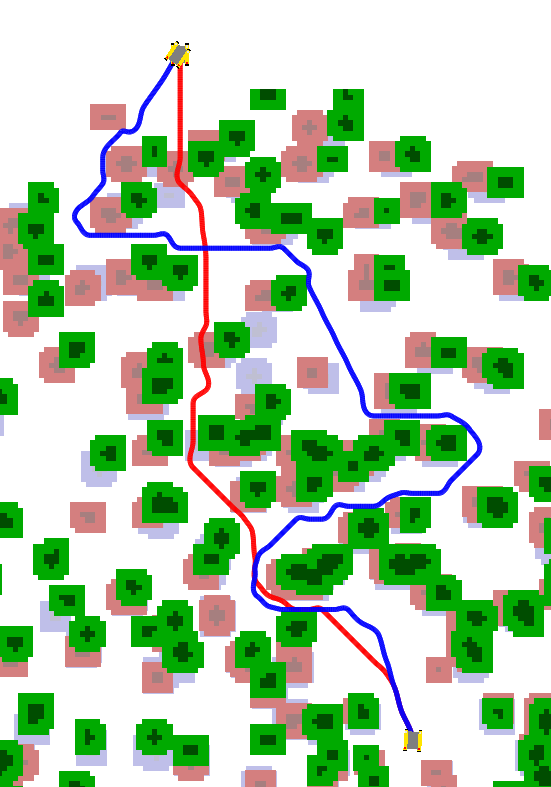}};
	\end{tikzpicture}
	\caption{The solution generated by VEH (blue) avoids all obstacles in all three hypothesized worlds. The solution from SH (red) only accounts for the most recent world hypothesis (green cost map).}
	\label{fig:VEH}
\vspace{-.4cm}
\end{figure}
Figure \ref{fig:VEH} shows the result of this method, where the trajectory from VEH (blue) is far more conservative than the trajectory generated when using only the most recent world hypothesis (red).
This occurs because VEH is unable to generate motions through areas of the map that were previously determined to have obstacles present.

\vspace{-.1cm}
\subsection{Per-Edge Hypothesis (PEH)}
To overcome the highly cautious planning of VEH while also considering safety, PEH was developed to compute edge costs that reflect the potential risk of incorrect assumptions without avoiding every obstacle in every world.
To do this, we implement a sub-search process into the search space, herein referred to as a Rerouter.
The Rerouter generates a trajectory from one position to another as in normal planning, but this trajectory is assigned as an edge between nodes where applicable.
We also allow expansions to occur if they are valid in at least one hypothetical world, although the final solution must be valid in the primary hypothesis.
This constraint is satisfied naturally when the final solution is extracted from the node history corresponding to the primary hypothesis.
Pseudocode for PEH is shown in Algorithm \ref{alg:PEH}.

\begin{algorithm}
\footnotesize
\caption{PEH}\label{alg:PEH}
\begin{algorithmic}[1]
\STATE $OPEN \gets$ {start node}
\STATE $CLOSED \gets \emptyset$
\WHILE{$OPEN$ not empty}
    \STATE $n \gets$ $OPEN$.pop()
		\STATE $CLOSED \gets n$
    \IF{$n$ is goal}
        \RETURN extractSolution($n$)
    \ENDIF
    \FORALL{$n'$ in Expand($n$)}
        \IF{$n'$ valid in any hypothesis}
            \FORALL{hypotheses where $n \to n'$ invalid}
                \STATE $n' \gets$ Reroute($n$, $n'$, hypothesis)
            \ENDFOR
            \STATE UpdateCost($n$, $n'$)
            \STATE $OPEN$.insert($n'$)
        \ENDIF
    \ENDFOR
\ENDWHILE
\end{algorithmic}
\end{algorithm}

\begin{figure}
	\vspace{-.3cm}
	\centering
	\noindent\resizebox{.9\linewidth}{!}{\begin{tikzpicture}
	\def\s{3.5}

	\begin{scope}[shift={(1.35*\s,0*\s)}]
		\fill[pattern=checkerboard, pattern color=white!90!black] (\s*.4,\s*.7) rectangle ++(\s*.2,\s*.6);
		\draw[very thick] (\s*0,\s*1) -- (\s*1, \s*1);
		\coordinate (a) at (\s*0,\s*1);
		\coordinate (aa) at (\s*.5,\s*1.5);
		\coordinate (b) at (\s*1,\s*1);
		\draw[dashed,very thick] (a) to[out=0,in=180] (aa) to[out=0,in=180] (b);

		\node[circle, draw, text centered, fill=white] at (\s*0,\s*1) {$n_{0,0}$}; 
		\node[circle, draw, text centered, fill=white] at (\s*1,\s*1) {$n_{1,0}$}; 
		\fill[pattern=checkerboard, pattern color=white!90!black] (-2.0,5.75) rectangle ++(1,.5);

	\end{scope}
	\begin{scope}[shift={(.25*\s,1.7*\s)}]
		\node[circle, draw, minimum size=.5cm,label=Node,fill=white] at (0,.035) {};
		\node at (2.4,.54) {Inconsistent Obstacle}; 
		\draw[very thick] (5.75,0) to (6.75,0);
		\node[label=Hypothesis 1 Expansion] at (5.9,.1) {}; 
		\draw[dashed,very thick] (9.25,0) to (10.25,0);
		\node[label=Hypothesis 2 Expansion] at (9.65,.1) {}; 
	\end{scope}
\end{tikzpicture}}
	\caption{Case 1: An inconsistent obstacle (gray checkerboard) does not exist in the primary world hypothesis, but does in the secondary world hypothesis. It falls within one edge expansion between nodes $n_{0,0}$ and $n_{1,0}$. The ``Hypothesis 1'' expansion assumes the region is free-space, and the ``Hypothesis 2'' expansion assumes the region is lethal.
	}
	\label{fig:bfMultiverse1}
\end{figure}
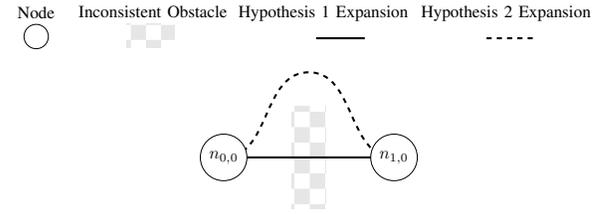

Figure \ref{fig:bfMultiverse1} shows an example where a hypothetical obstacle exists between nodes $n_{0,0}$ and $n_{1,0}$.
In the primary hypothesis where the obstacle does not exist, an edge is generated through the free space connecting the nodes.
In the secondary hypothesis where the obstacle does exist, the Rerouter generates a trajectory connecting the nodes, and the cost of $n_{1,0}$ is updated as the average of the two hypothetical expansions.
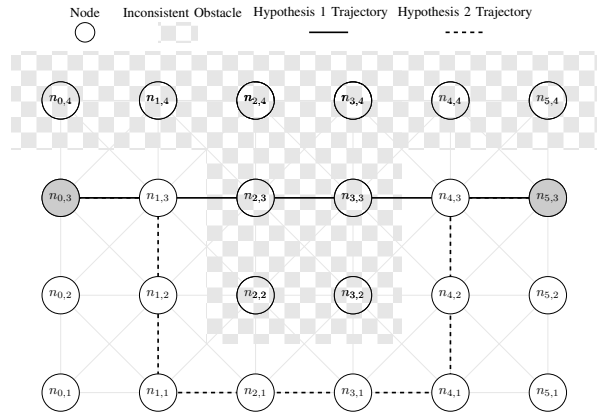
\begin{figure}[H]
	\vspace{-.3cm}
	\centering
	\noindent\resizebox{.9\linewidth}{!}{\begin{tikzpicture}
	\def\s{2.5}
	\fill[pattern=checkerboard, pattern color=white!90!black] (\s*-.5,\s*3.5) rectangle ++(\s*6,\s*1);
	\fill[pattern=checkerboard, pattern color=white!90!black] (\s*1.5,\s*1.5) rectangle ++(\s*2,\s*3);
	\foreach \x in {0,...,5}
		\foreach \y in {1,...,3}
			\draw[thin,color=white!90!black] (\s*\x,\s*\y) to (\s*\x,\s*\y+\s*1);
	\foreach \x in {0,...,4}
		\foreach \y in {1,...,4}
			\draw[thin,color=white!90!black] (\s*\x,\s*\y) to (\s*\x+\s*1,\s*\y);
	\foreach \x in {0,...,4}
		\foreach \y in {1,...,3}
			\draw[thin,color=white!90!black] (\s*\x,\s*\y) to (\s*\x+\s*1,\s*\y+\s*1);
	\foreach \x in {0,...,4}
		\foreach \y in {2,...,4}
			\draw[thin,color=white!90!black] (\s*\x,\s*\y) to (\s*\x+\s*1,\s*\y-\s*1);
	\draw[very thick] (\s*0,\s*3) -- (\s*5, \s*3);
	\coordinate (a) at (\s*0,\s*3);
	\coordinate (aa) at (\s*1,\s*3);
	\coordinate (b) at (\s*1,\s*2);
	\coordinate (c) at (\s*1,\s*1);
	\coordinate (d) at (\s*2,\s*1);
	\coordinate (e) at (\s*3,\s*1);
	\coordinate (f) at (\s*4,\s*1);
	\coordinate (g) at (\s*4,\s*2);
	\coordinate (gg) at (\s*4,\s*3);
	\coordinate (h) at (\s*5,\s*3);
	\draw[dashed,very thick] (a) to[out=0,in=180] (aa)
															 to[out=-90,in=90] (b)
															 to[out=-90,in=90] (c)
															 to[out=0,in=180] (d)
															 to[out=0,in=180] (e)
															 to[out=0,in=180] (f)
															 to[out=90,in=-90] (g)
															 to[out=90,in=-90] (gg)
															 to[out=0,in=-180] (h);
	\foreach \x in {0,...,5}
		\foreach \y in {1,...,4}
		\node[circle, draw, text centered,fill=white] at (\s*\x,\s*\y) {$n_{\x,\y}$}; 
	\foreach \x in {2,...,3}
		\foreach \y in {2,...,4}
		\node[circle, draw, text centered,pattern=checkerboard, pattern color=white!90!black] at (\s*\x,\s*\y) {$n_{\x,\y}$}; 
	\foreach \x in {0,...,5}
		\node[circle, draw, text centered,pattern=checkerboard, pattern color=white!90!black] at (\s*\x,\s*4) {$n_{\x,4}$}; 
	\node[circle, draw, text centered, fill=white!80!black] at (\s*0,\s*3) {$n_{0,3}$}; 
	\node[circle, draw, text centered, fill=white!80!black] at (\s*5,\s*3) {$n_{5,3}$}; 
	\fill[pattern=checkerboard, pattern color=white!90!black] (\s*1.0,\s*4.6) rectangle ++(1,.5);

	\begin{scope}[shift={(.25*\s,4.7*\s)}]
		\node[circle, draw, minimum size=.5cm,label=Node,fill=white] at (0,0) {}; 
		\node at (2.5,.5) {Inconsistent Obstacle}; 
		\draw[very thick] (5.75,0) to (6.75,0);
		\node[label=Hypothesis 1 Trajectory] at (6.05,.1) {}; 
		\draw[dashed,very thick] (9.25,0) to (10.25,0);
		\node[label=Hypothesis 2 Trajectory] at (9.75,.1) {}; 
	\end{scope}
\end{tikzpicture}}
	\caption{Case 2: An inconsistent obstacle (gray checkerboard) does not exist in the primary world hypothesis and does in the secondary world hypothesis. The ``Hypothesis 1'' and ``Hypothesis 2'' trajectories generated between nodes $n_{0,3}$ and $n_{5,3}$ assume the region is safe and lethal respectively.
	}
	\label{fig:bfMultiverse2}
\end{figure}

Figure \ref{fig:bfMultiverse2} shows the second case, where a hypothetical obstacle exists through multiple node expansions.
In the first case, rerouting between successive nodes is comparatively simple, but in this scenario, the Rerouter must be aware of the history of expansions for each node in every hypothesis.
This is because there is no way for an edge to be generated from $n_{1,3}$ to $n_{2,3}$ or $n_{3,3}$ in the secondary hypothesis.
It is not until an expansion is performed to $n_{4,3}$ that the Rerouter can look back through the node expansion history and generate an edge around the hypothetical obstacle between $n_{1,3}$ and $n_{4,3}$.

In theory, this methodology of PEH is effective in updating edge costs and maintaining consistency across hypotheses.
Although the rerouted solution costs are guaranteed to be no greater than those generated by VEH (assuming optimal sub-search and infinite planning time), the added computational cost associated with performing reroutes for every edge rendered PEH impractical for our field experiments.

\subsection{Goal-Edge Hypothesis (GEH)}
To overcome the long planning times of PEH, we developed GEH which defers the notion of multiple hypothetical worlds until after an edge expands into the goal.
At that point, the Rerouter generates an edge from the first point of divergence in each hypothesis to the goal.
The goal-edge cost is updated as the average of the new rerouted trajectories, the node is reinserted into the open list, and search continues.
The final trajectory may not fully avoid obstacles in non-primary hypotheses, but it remains conservative, favoring motions that pass through the edges of obstacles rather than cutting straight through their centers.
Pseudocode for GEH is shown in Algorithm \ref{alg:GEH}.
\begin{algorithm}
\footnotesize
\caption{GEH}\label{alg:GEH}
\begin{algorithmic}[1]
	\STATE $OPEN \gets$ {start node}
	\STATE $CLOSED \gets \emptyset$
	\WHILE{$OPEN$ not empty}
			\STATE $n \gets$ $OPEN$.pop()
			\STATE $CLOSED \gets n$
			\IF{$n$ is goal}
					\FORALL{hypotheses}
							\IF{path invalid}
									\STATE Reroute from divergence point to goal
									\STATE Update goal-edge cost
							\ENDIF
					\ENDFOR
					\STATE Reinsert goal node into $OPEN$
					\STATE \textbf{continue}
			\ENDIF
			\FORALL{$n'$ in Expand($n$)}
					\IF{$n'$ valid in any hypothesis}
							\STATE UpdateCost($n$, $n'$)
							\STATE $OPEN$.insert($n'$)
					\ENDIF
			\ENDFOR
	\ENDWHILE
\end{algorithmic}
\end{algorithm}

\vspace{-.1cm}
\begin{figure}
	\centering
	\noindent\resizebox{.95\linewidth}{!}{\begin{tikzpicture}
	\def\s{2.5}
	\fill[pattern=checkerboard, pattern color=white!90!black] (\s*-.5,\s*3.5) rectangle ++(\s*6,\s*1);
	\fill[pattern=checkerboard, pattern color=white!90!black] (\s*1.5,\s*1.5) rectangle ++(\s*2,\s*3);
	\foreach \x in {0,...,5}
		\foreach \y in {1,...,3}
			\draw[thin,color=white!90!black] (\s*\x,\s*\y) to (\s*\x,\s*\y+\s*1);
	\foreach \x in {0,...,4}
		\foreach \y in {1,...,4}
			\draw[thin,color=white!90!black] (\s*\x,\s*\y) to (\s*\x+\s*1,\s*\y);
	\foreach \x in {0,...,4}
		\foreach \y in {1,...,3}
			\draw[thin,color=white!90!black] (\s*\x,\s*\y) to (\s*\x+\s*1,\s*\y+\s*1);
	\foreach \x in {0,...,4}
		\foreach \y in {2,...,4}
			\draw[thin,color=white!90!black] (\s*\x,\s*\y) to (\s*\x+\s*1,\s*\y-\s*1);
	\draw[very thick] (\s*0,\s*3) -- (\s*5, \s*3);
	\coordinate (a) at (\s*0,\s*3);
	\coordinate (aa) at (\s*1,\s*3);
	\coordinate (b) at (\s*1,\s*2);
	\coordinate (c) at (\s*1,\s*1);
	\coordinate (d) at (\s*2,\s*1);
	\coordinate (e) at (\s*3,\s*1);
	\coordinate (f) at (\s*4,\s*1);
	\coordinate (g) at (\s*4,\s*2);
	\coordinate (h) at (\s*5,\s*3);
	\coordinate (i) at (\s*5,\s*2);
	\coordinate (j) at (\s*2,\s*2);
	\coordinate (k) at (\s*3,\s*2);
	\draw[dashed,very thick] (a) to[out=0,in=180] (aa)
															 to[out=-90,in=90] (b)
															 to[out=-90,in=90] (c)
															 to[out=0,in=180] (d)
															 to[out=0,in=180] (e)
															 to[out=0,in=180] (f)
															 to[out=45,in=-135] (i)
															 to[out=90,in=-90] (h);
	\draw[dotted,very thick] (a) to[out=-45,in=135] (b)
															 to[out=0,in=-180] (j)
															 to[out=0,in=-180] (k)
															 to[out=0,in=-180] (g)
															 to[out=45,in=-135] (h);

	\foreach \x in {0,...,5}
		\foreach \y in {1,...,4}
			\node[circle, draw, text centered,fill=white] at (\s*\x,\s*\y) {$n_{\x,\y}$}; 
	\foreach \x in {2,...,3}
		\foreach \y in {2,...,4}
			\node[circle, draw, text centered,pattern=checkerboard, pattern color=white!90!black] at (\s*\x,\s*\y) {$n_{\x,\y}$}; 
	\foreach \x in {0,...,5}
		\node[circle, draw, text centered,pattern=checkerboard, pattern color=white!90!black] at (\s*\x,\s*4) {$n_{\x,4}$}; 

	\node[circle, draw, text centered, fill=white!80!black] at (\s*0,\s*3) {$n_{0,3}$}; 
	\node[circle, draw, text centered, fill=white!80!black] at (\s*5,\s*3) {$n_{5,3}$}; 
	\fill[pattern=checkerboard, pattern color=white!90!black] (\s*.55,\s*4.6) rectangle ++(1,.5);
	\begin{scope}[shift={(-.25*\s,4.7*\s)}]
		\node[circle, draw, minimum size=.5cm,label=Node,fill=white] at (0,0) {}; 
		\node at (2.5,.5) {Inconsistent Obstacle}; 
		\draw[very thick] (5.75,0) to (6.75,0);
		\node[label=Hypothesis 1 Trajectory] at (6.05,.1) {}; 
		\draw[dashed,very thick] (9.25,0) to (10.25,0);
		\node[label=Hypothesis 2 Trajectory] at (9.75,.1) {}; 
		\draw[dotted,very thick] (12.25,0) to (13.25,0);
		\node[label=Final Trajectory] at (12.85,.1) {}; 
	\end{scope}
\end{tikzpicture}}
	\caption{An inconsistent obstacle (gray checkerboard) exists between the start and goal states (dark gray nodes). A route (``Hypothesis 1 Trajectory'') is found directly to the goal because the expansions are valid in at least one hypothesis - in this case, the primary. The final trajectory is a result of considering the cost of rerouting in the secondary hypothesis.
	}
	\label{fig:dfMultiverse1}
\end{figure}
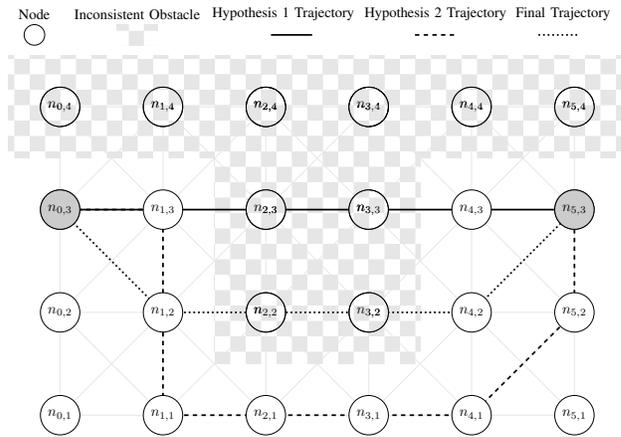

Figure \ref{fig:dfMultiverse1} depicts the process, where the final trajectory is influenced by adjusting the cost of the expansion from $n_{4,3}$ to $n_{5,3}$ with the rerouted edge from $n_{1,3}$ to the goal, $n_{5,3}$.
After $n_{4,3}$ expands to the goal, a reroute is performed from the node corresponding to the first divergence point ($n_{1,3}$), generating the ``Hypothesis 2 trajectory''. The cost of the expansion from $n_{4,3}$ to $n_{5,3}$ is updated by averaging the original primary hypothesis expansion to the goal with the rerouted trajectory to the goal. After search is finished, the resulting ``Final Trajectory'' is one that expands $n_{0,3}$ to $n_{1,2}$, and cuts through the very edge of the obstacle in the secondary hypothesis.
When search is complete, the final trajectory balances between going through the center of the uncertain area and completely avoiding it.

\begin{figure}
	\centering
	\begin{tikzpicture}
		\node[anchor=south west,inner sep=0] at (0,0) {\includegraphics[width=.95\linewidth]{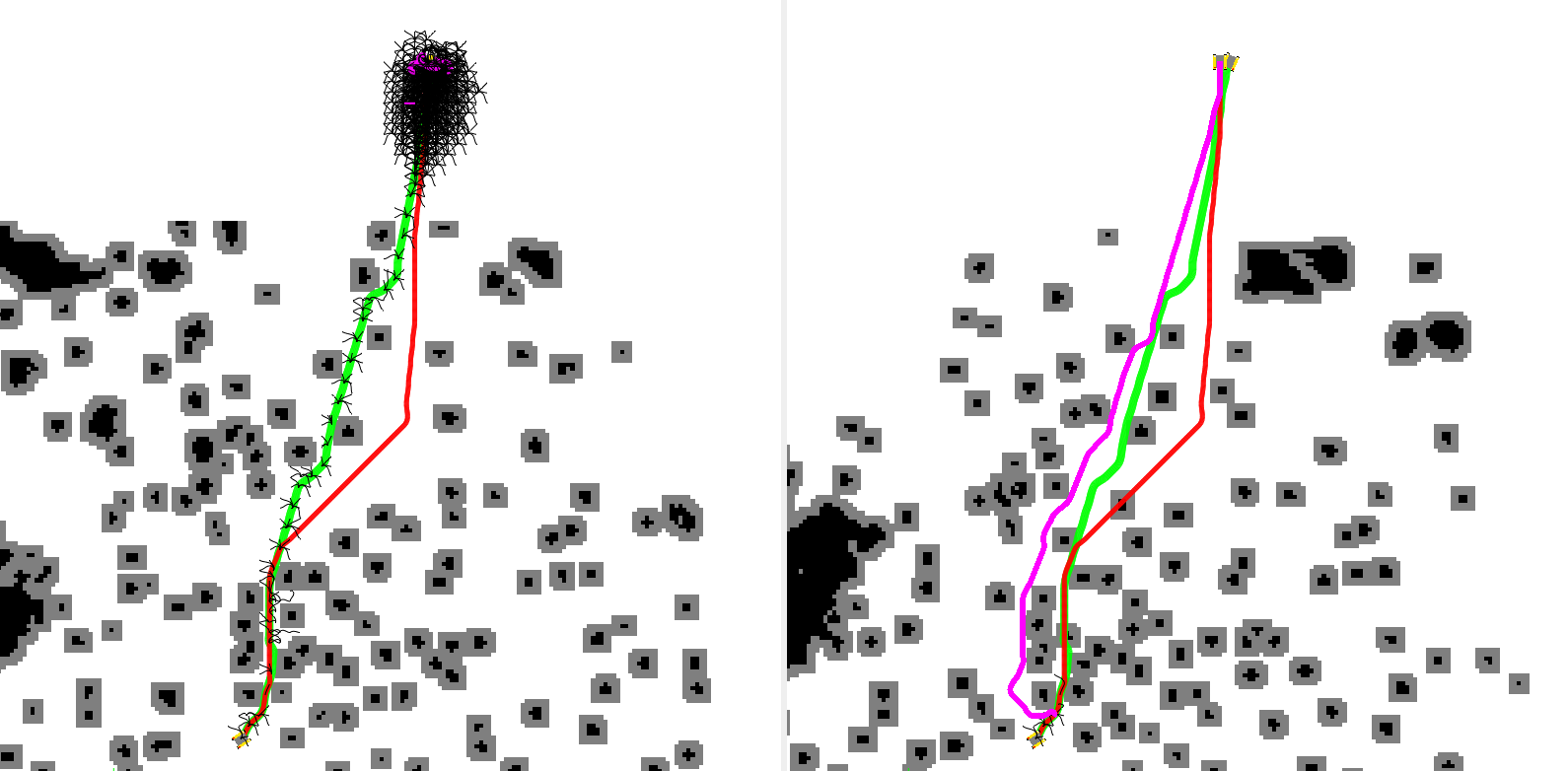}};
		\node[font=\scriptsize] at (1.1,3.85) {Primary Hypothesis};
		\node[font=\scriptsize] at (5.5,3.85) {Secondary Hypothesis};
	\end{tikzpicture}
	\caption{GEH search space from field data for a primary and secondary hypothetical. Expansions are shown in black. GEH final solution is shown in green, the primary-hypothesis solution is shown in red, and the secondary-hypothesis reroute is shown in magenta in the map on the right. }
	\label{fig:GEHSearchSpace}
\end{figure}

Figure \ref{fig:GEHSearchSpace} illustrates the result of this process with an example from the field data, where the GEH solution is a compromise between the primary-hypothesis solution and the secondary-hypothesis rerouted solution.
This figure also demonstrates the primary issue with this method, particularly for long-range planning.
Because the cost of the goal-expanded edge is the only one updated by the Rerouter, graph expansions are concentrated at the final state as search continually updates and reinserts the candidate terminal node back into the open list.
This effect is magnified with the inflated heuristic of ARA$^*$ because edges closer to the goal are more likely to be expanded than those further away.
GEH is more likely to succeed if divergence points only occur near the goal, or if the goal is close to the robot starting state.
It is important to note that the final plan generated by GEH is a result of multiple consistent trajectory branches, where each world hypothesis may diverge from the primary solution at different points in the graph.
Rather than a single common trajectory across all hypotheses, GEH maintains $n$ partially consistent paths, each used to update node costs and balance obstacle avoidance with exploration in uncertain regions.

\subsection{Goal-Edge Graph Revision Hypothesis (GEGRH)}
This final method solves the goal-expansion issue in GEH by including a graph revision step after updating the goal-expansion cost. 
The first step of the revision is to step back through the history of node expansions and determine the node corresponding to the first divergence point of all hypothetical worlds.
This is performed using the restructured node-history described in Section \ref{sec:searchSpace}.
Following the divergence point identification, the graph revision steps are shown in Algorithm \ref{alg:graphRevision}.
As in normal A$^*$ search, if the reinserted node has the lowest f-cost on the open list, search is complete, or as in ARA$^*$, the heuristic is deflated and search resumes.
This allows the graph to retain prior candidate solutions as independent edges and focus expansions at the divergence points of the hypotheses.
Pseudocode for GEGRH is shown in Algorithm \ref{alg:GEGRH}. 
\vspace{-.2cm}
\begin{algorithm}
\footnotesize
\caption{Graph Revision}\label{alg:graphRevision}
\begin{algorithmic}[1]
	\STATE $n_e$ = goal expanded node
	\STATE $n_d$ = divergence node
	\STATE $n_e$.backPointer $=$ {$n_d$.backPointer}
	\STATE $n_d$.RemoveNodeDescendents($OPEN$,$CLOSED$)
	\STATE $OPEN \gets n_e$
	\STATE $n_d$.MarkInvalid()
\end{algorithmic}
\end{algorithm}

\vspace{-.9cm}
\begin{algorithm}
\footnotesize
\caption{GEGRH}\label{alg:GEGRH}
\begin{algorithmic}[1]
	\STATE $OPEN \gets$ {start node}
	\STATE $CLOSED \gets \emptyset$
	\WHILE{$OPEN$ not empty}
			\STATE $n \gets$ $OPEN$.pop()
			\STATE $CLOSED \gets n$
			\IF{$n$ is goal}
					\FORALL{hypotheses}
							\IF{path invalid}
									\STATE Reroute from divergence point to goal
									\STATE Update goal-edge cost
							\ENDIF
					\ENDFOR
					\STATE Perform the graph revision step 
					\STATE \textbf{continue}
			\ENDIF
			\FORALL{$n'$ in Expand($n$)}
					\IF{$n'$ valid in any hypothesis}
							\STATE UpdateCost($n$, $n'$)
							\STATE $OPEN$.insert($n'$)
					\ENDIF
			\ENDFOR
	\ENDWHILE
\end{algorithmic}
\end{algorithm}

\vspace{-.7cm}
\begin{figure}[H]
	\centering
	\begin{tikzpicture}
		\node[anchor=south west,inner sep=0] at (0,0) {\includegraphics[width=.95\linewidth]{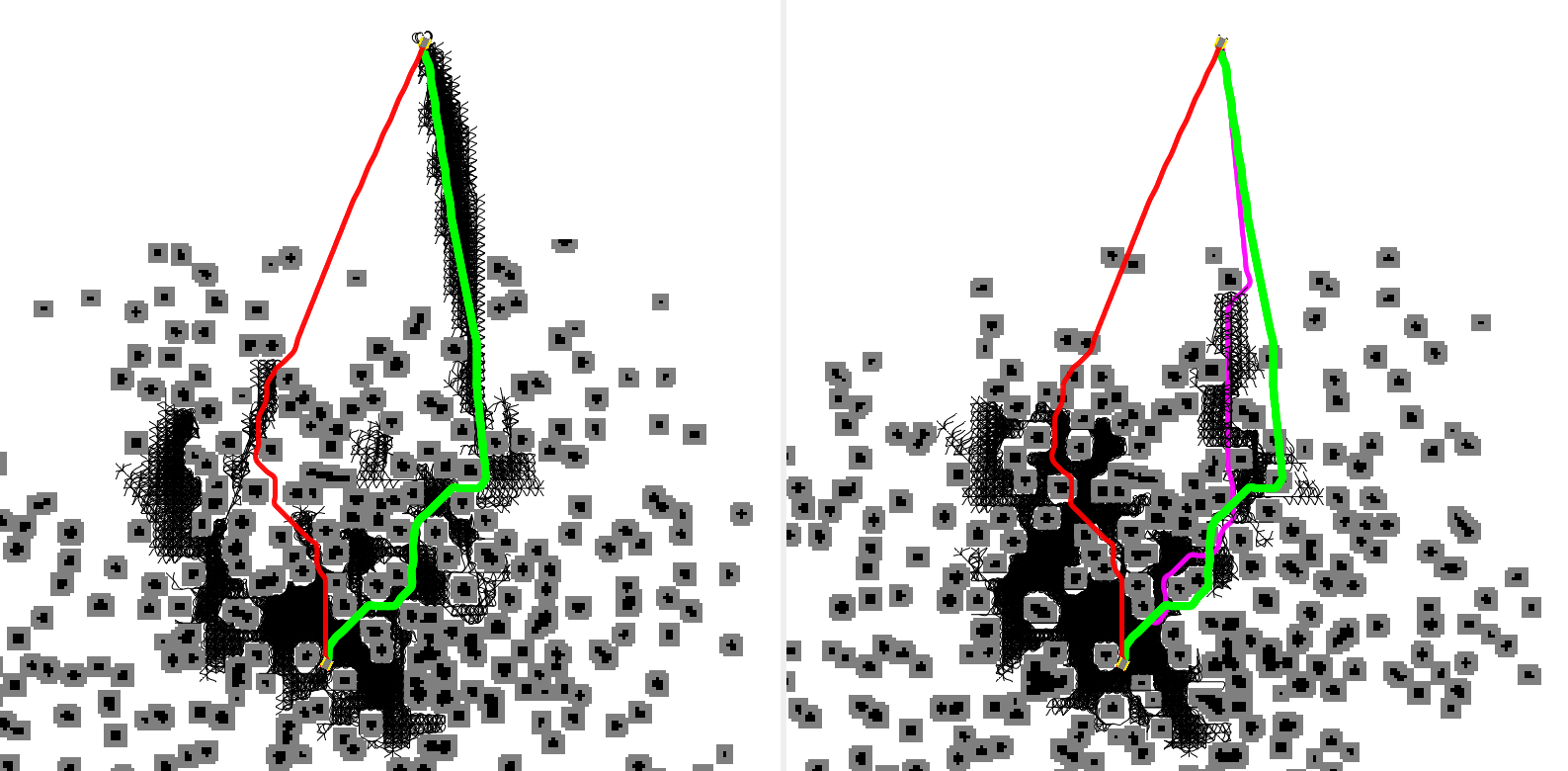}};
		\node[font=\scriptsize] at (1.1,3.85) {Primary Hypothesis};
		\node[font=\scriptsize] at (5.5,3.85) {Secondary Hypothesis};
	\end{tikzpicture}
	\caption{GEGRH search space from field test data for a primary and secondary hypothetical. Expansions shown in black. GEGRH final solution  shown in green, primary-hypothesis solution shown in red, and secondary-hypothesis reroute is shown in magenta in the map on the right.}
	\label{fig:GEGRHSearchSpace}
\end{figure}

Figure \ref{fig:GEGRHSearchSpace} shows the resulting search space after adding the graph revision step.
Note the difference in where the graph expansions are concentrated compared to in Figure \ref{fig:GEHSearchSpace}.
Focusing expansions on the divergence points rather than around the goal allows solutions to be found more effectively.

GEGRH is also able to find solutions when GEH should perform well.
In theory, GEH has marginally less computing overhead when the divergence point is very close the goal, but the efficiency increases of GEGRH throughout the rest of search minimize the impact of these cases.
\section{Experimental Design}
To evaluate the impact of different methods for incorporating multiple hypothetical versions of the world into KEASL, experiments were conducted using data collected from testing on a UGV.
During testing, environment maps were generated online by the perception system. These maps, along with recorded start and goal states, formed the basis of 221 planning problems.
For each planning problem, two experimental conditions were tested:
\begin{enumerate}
    \item Planning with two successive environment maps.
    \item Planning with three successive environment maps.
\end{enumerate}
In all experiments, ARA$^*$ was used as the underlying search algorithm, with a maximum allowable planning time of 1-second, and an initial heuristic inflation factor of 2.0.
Each planning problem was solved using three planning modes - SH, VEH, and GEGRH.
For each version of the search space, we recorded both the planning time and the resulting trajectory duration.
We focus our analysis on VEH and GEGRH, as preliminary evaluations indicated that PEH incurred impractical runtimes for field use, and that GEH was superseded by GEGRH. SH is included for context of solutions generated by normal search.
VEH is considered the baseline because it represents the most conservative method of planning through multiple maps and uses the same KEASL framework to enable a fair comparison of the algorithmic effects.

\begin{figure}[htb]
	\centering
	\begin{subfigure}{0.3\linewidth}
		\includegraphics[width=\linewidth]{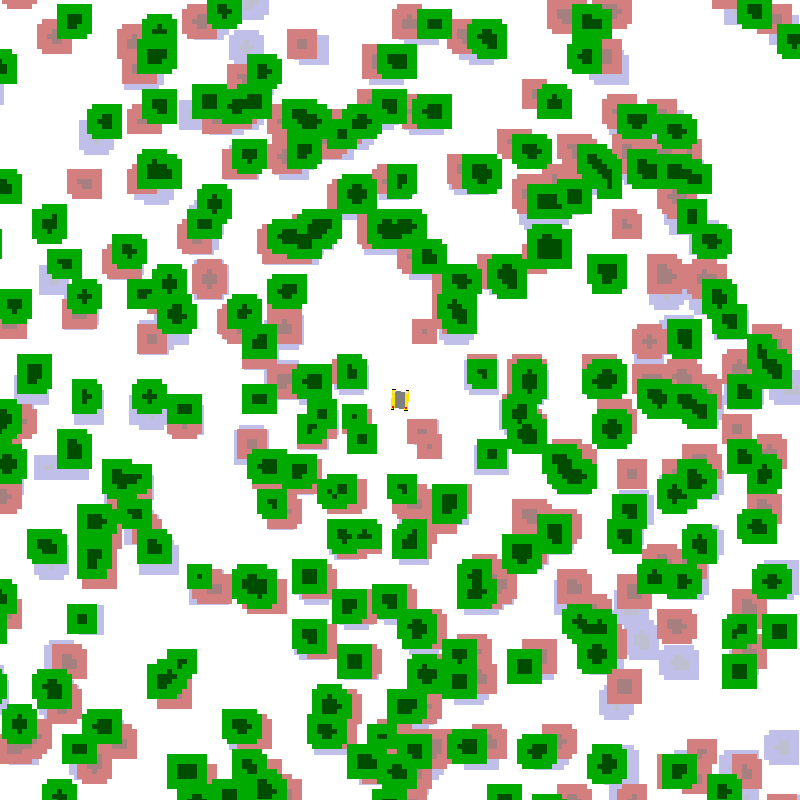}
		\caption{}
	\end{subfigure}
	\begin{subfigure}{0.3\linewidth}
		\includegraphics[width=\linewidth]{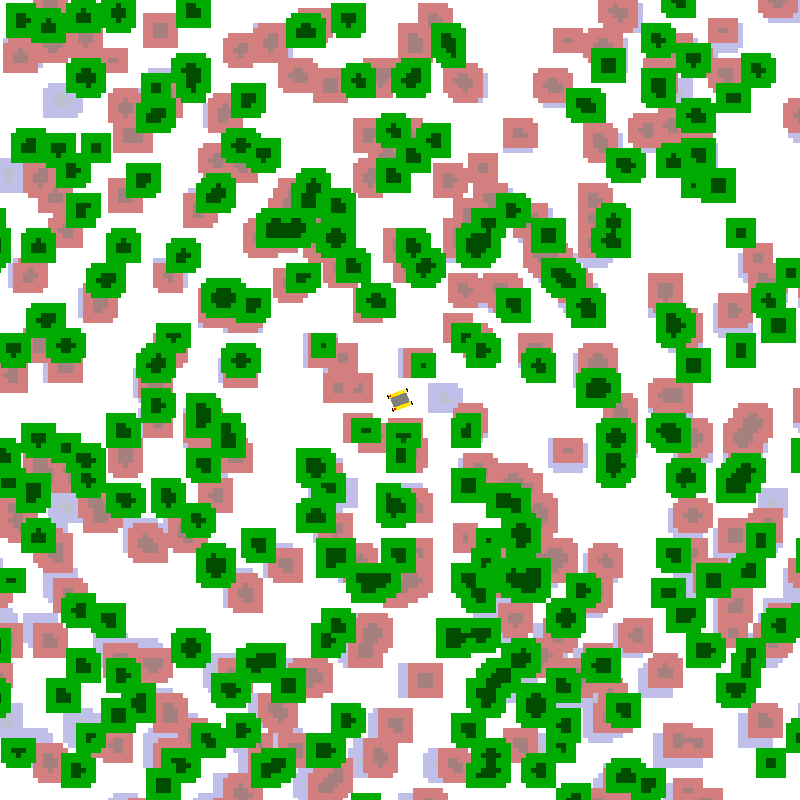}
		\caption{}
	\end{subfigure}
	\begin{subfigure}{0.3\linewidth}
		\includegraphics[width=\linewidth]{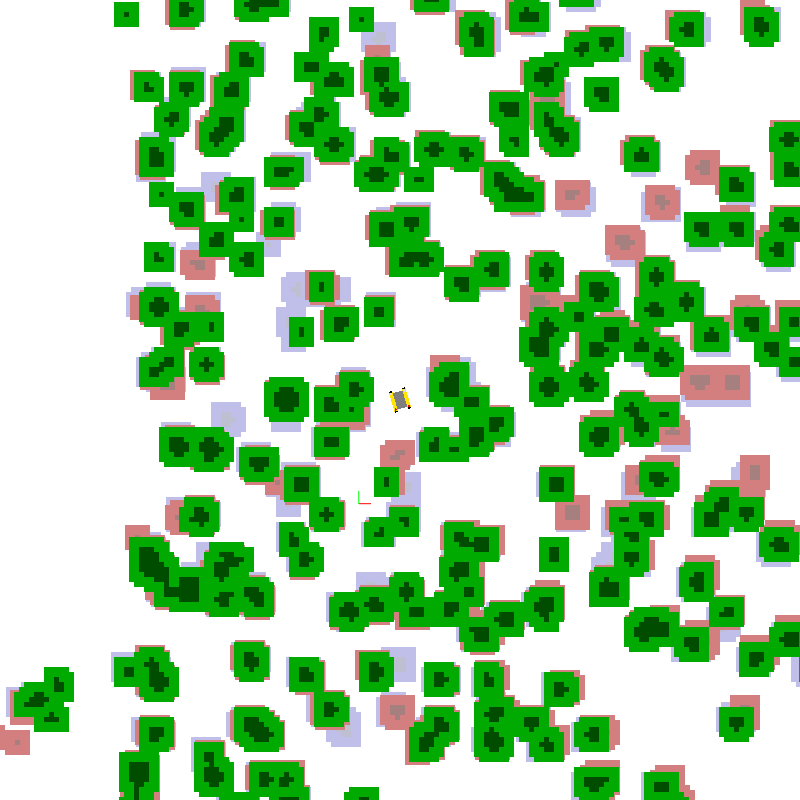}
		\caption{}
	\end{subfigure}
	\caption{Sample of three planning problem worlds from the experiments, with the primary hypothesis (green) overlaid on the secondary hypothesis (red), overlaid on the tertiary hypothesis (blue). Visual fading is applied between layers for contrast.}
	\label{fig:threeOverlaidWorlds}
\end{figure}

An example of the experiment representations is shown in Figure \ref{fig:threeOverlaidWorlds}, where inconsistencies between successive environment maps can be observed despite high confidence levels from the perception and state estimation systems.
These inconsistencies come because of updated observations, as well as errors in state estimation which cause obstacles to appear slightly shifted or rotated in successive maps.
\section{Results \& Discussion}
\begin{table}[htb]
	\centering
	\begin{tabular}{p{2.2cm} || >{\centering\arraybackslash}p{1.5cm} | >{\centering\arraybackslash}p{1.5cm} | >{\centering\arraybackslash}p{1.5cm} ||}
		\multicolumn{4}{c}{Two Hypotheses}\\
		\hline
		Average & SH & VEH & GEGRH\\
		\hline\hline
		Planning Time (s) & $0.12\pm0.01$ & $0.26\pm0.03$ & $0.22\pm0.02$ \\
		Path Duration (s) & $33.55\pm1.70$ & $38.56\pm2.06$ & $36.00\pm1.87$ \\
		\hline
		\multicolumn{4}{c}{}\\
		\multicolumn{4}{c}{Three Hypotheses}\\
		\hline
		Average & SH & VEH & GEGRH\\
		\hline\hline
		Planning Time (s) & $0.12\pm0.01$ & $0.41\pm0.04$ & $0.29\pm0.03$ \\
		Path Duration (s) & $33.55\pm1.70$ & $44.87\pm2.91$ & $36.47\pm1.98$ \\
		\hline
	\end{tabular}
	\caption{Planning performance of VEH, GEGRH, and SH across 221 planning problems. Average values are shown with a 95\% confidence interval. SH is included as context to planning without multiple world models.}
	\label{table:MEASLdata}
\end{table}
The experiment shows the performance of GEGRH compared to VEH, with added context from SH planning.
Table \ref{table:MEASLdata} shows the average planning times and path durations of the methods based on a 95\% confidence interval.
GEGRH showed a 15.4\% (.04-second) decrease in planning time for two-hypothesis planning and a 29.27\% (.12-second) decrease in planning time for three-hypothesis planning compared to VEH.
This is likely due to GEGRH allowing search through areas that are not valid in all hypotheses, causing search to reach the goal faster.
The path duration was also improved by GEGRH when compared to VEH, which saw a 6.64\% (2.6-second) decrease for two-hypothesis and an 18.72\% (8.4-second) decrease for three-hypothesis.
This is because VEH must plan through the sum of all environment maps, and the graph becomes more difficult to navigate with additional worlds.
The gap in performance between GEGRH and VEH widens with the addition of one hypothetical world, and exploring the impact on this gap of adding more hypotheses is a research task of future interest.
The performance of SH is included for context of normal KEASL behavior.
Both VEH and GEGRH have higher planning times, which is expected because of the additional computation involved in processing multiple environment maps.
Additionally, the increased path durations indicate more conservative trajectory generation from the consideration of prior perception output.

\begin{figure}[htb]
	\centering
	\includegraphics[width=.95\linewidth]{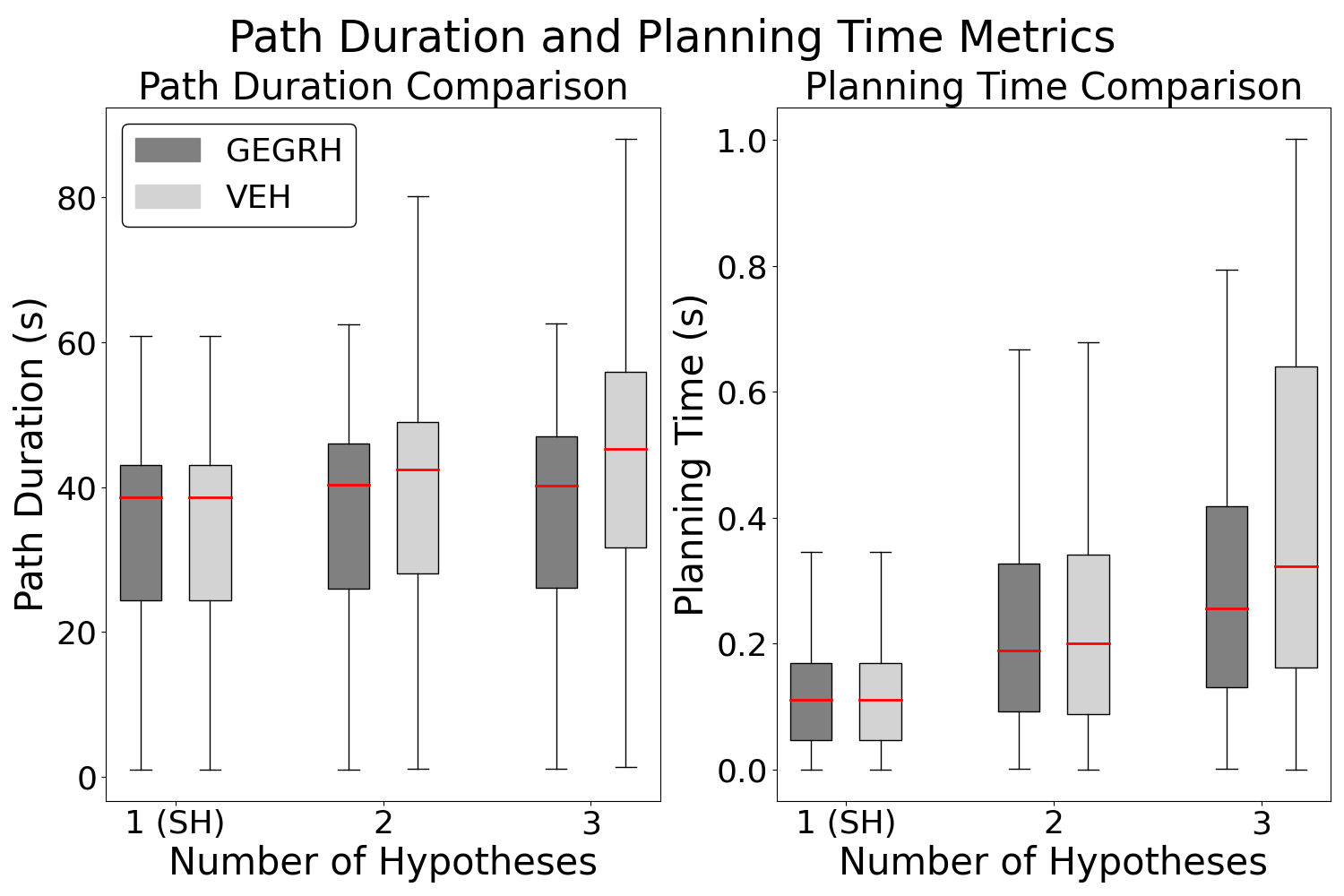}
	\caption{Path duration and planning time data for VEH and GEGRH from the field experiment data. \textbf{Gray boxes:} 25th to 75th percentile, \textbf{Red lines:} Median value (50th percentile), \textbf{Whiskers:} Data range (excluding outliers).}
	\label{fig:pathDurationMetrics}
\end{figure}
%

%

Figure \ref{fig:pathDurationMetrics} illustrates the path duration and planning time metrics of GEGRH, and VEH.
They perform the same when operating with one map, because they both default to SH planning.
The data ranges are similar for 2-hypothesis planning, with GEGR slightly outperforming VEH based on the median values for both path duration and planning time.
The gap widens with 3-hypothesis planning, likely due to the highly cluttered maps that VEH must plan through.
The range of VEH planning time data extends to one second, indicating a substantial number of planning cycles that take the full time constraint during ARA$^*$ search.

To give further context to the field experiments, we present two case studies which give qualitative comparisons of the outputs of VEH and GEGRH.
The first is one where GEGRH outperforms VEH and generates a more direct trajectory through the world.
In the second case, VEH outperforms GEGRH, highlighting points for possible future refinement.

\subsection{Case Study One}
\begin{figure}[htb]
	\centering
	\begin{tikzpicture}
		\node[anchor=south west,inner sep=0] at (0,0) {\includegraphics[width=.8\linewidth]{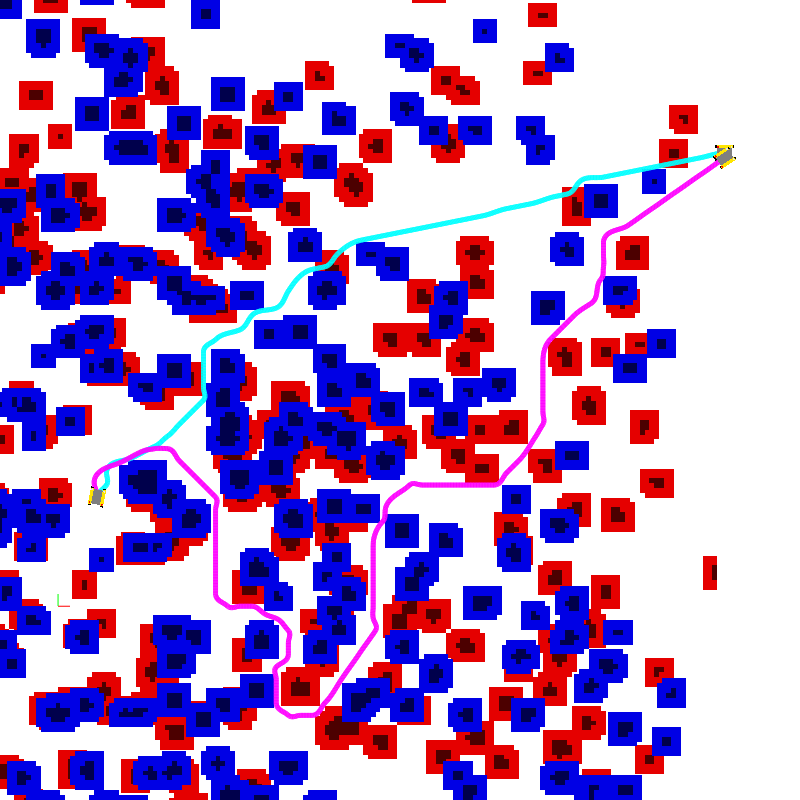}};
		\draw[rotate around={0:(1.35,1.2)}, green!80!black, very thick] (2.8,4.7) ellipse[x radius=0.7, y radius=0.4];
	\end{tikzpicture}
	\caption{ The first case study with solutions from GEGRH (cyan) and VEH (magenta) on a two-hypothesis map, with the primary hypothesis in blue and secondary in red. GEGRH produces a plan with lower cost than VEH. The region of interest is circled in green.}
	\label{fig:GEGRHBetter}
\end{figure}

This case study illustrated by Figure \ref{fig:GEGRHBetter} presents a planning problem where GEGRH can generate a solution with a lower traversal time than VEH.
In this case, there is a specific choke point in the map (circled in green in the figure) that VEH cannot expand through because of an obstacle present in the secondary hypothesis.
VEH plans a very cautious trajectory that avoids all obstacles in all world models and instead finds a route that is free of inconsistency.
Although there is a possibility of an obstacle existing at the choke point, GEGRH determines that the risk of needing to reroute around the choke point is worth exploring because of the potential decrease in path duration.
By tracking previous observations of the world in the search space, GEGRH can generate trajectories that are cautious of potential obstacles without being too optimistic about the true state of the world.

\subsection{Case Study Two}
\begin{figure}[htb]
	\centering
	\begin{tikzpicture}
		\node[anchor=south west,inner sep=0] at (0,0) { \includegraphics[width=.8\linewidth]{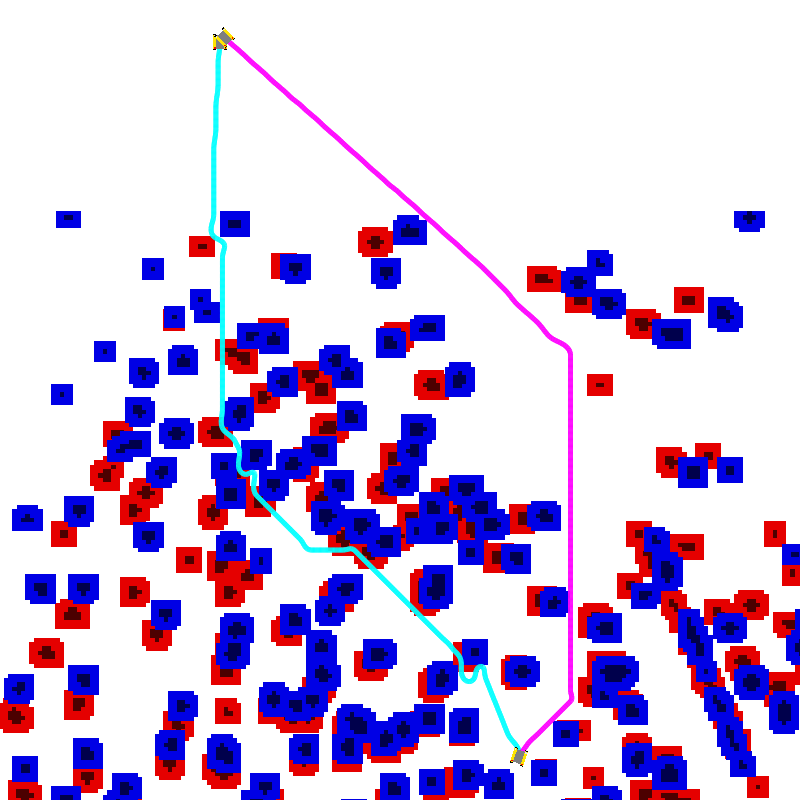} };
			\draw[rotate around={0:(1.0,5.0)}, green!80!black, very thick] (4.0,1.2) ellipse[x radius=0.7, y radius=0.4];
		\end{tikzpicture}
	\caption{The second case study with solutions from GEGRH (cyan) and VEH (magenta) on a two-hypothesis map, with the primary hypothesis in blue and secondary in red. VEH produces a plan with a lower cost than GEGRH. The area of interest is circled in green.}
	\label{fig:VEHBetter}
\end{figure}

The second case study shown in Figure \ref{fig:VEHBetter} shows a planning problem where VEH outperforms GEGRH by generating a trajectory with a lower traversal time.
This is because the rerouting sub-search procedure added too much time in this obstacle configuration and GEGRH was only able to find a suboptimal solution with an inflated heuristic within the one-second time constraint.
VEH likely generated a lower cost plan because it was forced to expand through the open area of the graph, which also led to a direct route to the goal.
The direct routing allowed it to find a path after deflating the heuristic during the ARA$^*$ search.
The winding motion present toward the bottom of the GEGRH solution (circled in green) is because of the shifted obstacle in the secondary hypothesis compared to the primary hypothesis.
In this case, the Rerouter in the primary hypothesis had to abruptly turn back and find a path around the obstacle.
Because it was unable to find an optimal solution within the one-second time limit (it was only able to find a suboptimal solution with an inflated heuristic), this path was the final solution because of the constraint that all final plans must be valid in the primary hypothesis.
\section{Conclusion, \& Future Considerations}
In environments where the world model is incomplete or uncertain, traditional single-hypothesis planning can lead to inconsistent plans that vary significantly under slight environmental changes.
By shifting from the assumption of uncertainty as hazardous to assigning edge costs that reflect potential risk of incorrect assumptions, planning can become more robust to inconsistencies in perception and state estimation.
The four methods explored in this work represent a progression in how to balance caution, computational efficiency, and adaptability to updated world information.
VEH guarantees validity across all hypotheses but can be overly conservative.
PEH relaxes this caution at the expense of high computational costs.
GEH improves efficiency by delaying constraint checking until a node expands to the goal, and GEGRH further improves with an additional graph-revision step.
These methods illustrate that it is possible to explicitly reason about perception uncertainty during deterministic graph search.

A future avenue of research is to increase the number of environment hypotheses incorporated into each search space, to better handle environments with frequent and severe perceptual inconsistencies.
Along with this, exploring ways to consider the divergence of non-binary map information like ground height or semantics is of high interest because of the complexity of the target environments.
Another potential direction is to explore the use of different cost functions when recalculating the $g$-cost of nodes after rerouting, rather than weighting the costs from all hypotheses evenly.
Additionally, incorporating an estimate of hypothesis certainty could allow a more nuanced adjustment of node costs and constraints, further improving planning efficiency and robustness.
\bibliographystyle{plain}
\bibliography{root}

\end{document}